\documentclass{isprs}
\usepackage{setspace}
\usepackage{geometry}
\usepackage{epstopdf}
\usepackage[labelsep=period]{caption} 
\usepackage[none]{hyphenat}
\usepackage{breakcites}
\usepackage[utf8]{inputenc}
\usepackage[T1]{fontenc}
\usepackage[utf8]{inputenc}
\usepackage{mathtools}
\usepackage{graphicx}
\usepackage{float}
\usepackage{blindtext}
\usepackage{amsmath}
\usepackage[vlined,ruled]{algorithm2e}
\usepackage{subfig}

\def\figurePath{figures/}

\begin{document}

\title{Weighted point cloud augmentation for neural network training data class-imbalance}

\author{
 David Griffiths\textsuperscript{a, }\thanks{Corresponding author.}
 , Jan Boehm\textsuperscript{a}}

\address
{
	\textsuperscript{a }Dept. of Civil, Environmental and Geomatic Engineering, University College London, Gower Street, London, \\ WC1E 6BT UK - (david.griffiths.16, j.boehm)@ucl.ac.uk
}

\commission{II, }{II}
\workinggroup{II/3}
\icwg{}

\abstract{
Recent developments in the field of deep learning for 3D data have demonstrated promising potential for end-to-end learning directly from point clouds. However, many real-world point clouds contain a large class im-balance due to the natural class im-balance observed in nature. For example, a 3D scan of an urban environment will consist mostly of road and fa{\c c}ade, whereas other objects such as poles will be under-represented. In this paper we address this issue by employing a weighted augmentation to increase classes that contain fewer points. By mitigating the class im-balance present in the data we demonstrate that a standard PointNet++ deep neural network can achieve higher performance at inference on validation data. This was observed as an increase of F1 score of 19\% and 25\% on two test benchmark datasets; ScanNet and Semantic3D respectively where no class im-balance pre-processing had been performed. Our networks performed better on both highly-represented and under-represented classes, which indicates that the network is learning more robust and meaningful features when the loss function is not overly exposed to only a few classes.
}

\maketitle
\sloppy
\section{Introduction}

The success of deep learning for 2D image processing has been due to a combination of improved hardware, software and data. Despite advances in 2D data processing, it is evident that progress in 3D data is still far behind \cite{HackelEtAlT2017}. Processing of 3D geometry such as point clouds can be performed using the same hardware and software libraries (i.e. tensorflow, torch, caffe) as 2D image processing, and recently there has been a surge in network architectures that can learn directly from point clouds in an end-to-end manner. Such examples include; PointNet \cite{CharlesEtAlR2017}, PointNet++ \cite{QiEtAlC2017a}, SPLATNet \cite{SuEtAlH2018}, PointCNN \cite{LiEtAlY2018} and MCCNN \cite{HermosillaEtAlP2018}. This is currently a very active area of research and offers exciting potential. 

Classic machine learning approaches for point cloud classification use hand-crafted feature descriptors, which are computed for each individual point. For such approaches each point represents one training sample. Even moderately sized labelled point clouds therefore are adequate for training in such a framework. In contrary, deep learning methods for per-point classification typically operate on small sub-sets or sub-windows of the point cloud representing a whole scene. Every sub-window represents one training sample. It is immediately clear that the number of training samples becomes an issue. However, the success of 2D deep learning is often largely accredited to the release of large open-access labelled datasets such as ImageNet \cite{DengJ2009}, which contains $> 14*10^6$ images. It is now largely standard procedure to pre-train deep CNNs on the ImageNet benchmark dataset for initial model weight tuning. Achieving a similar dataset for 3D point cloud processing would be a substantially more challenging feat, and as such open training datasets on the scale of ImageNet do not exist for 3D point clouds. Regardless, there has been a range of efforts to address this issue. The most obvious attempt for a 3D ImageNet comes in the form of ShapeNet \cite{ChangEtAlA2015a}. ShapeNet contains over 300 million models with 220,000 classified into 3,135 classes arranged using WordNet hypernym-hyponym relationships. Similarly, ScanNet \cite{DaiEtAlA2017} contains over 1500 indoor scene scans, with each scan containing ~400-600k points. With respect to outdoor point cloud processing there have also been significant efforts to address this problem, most noticeably; iQmumuls/TerraMobilita \cite{ValletEtAlB2015}, TUM City Campus \cite{GehrungEtAlJ2017} and the current largest, Semantic3D \cite{HackelEtAlT2017} which contains ~4 billion points.

Although these datasets offer large point counts in absolute terms, they contain very large class-imbalances. This is due to the natural class imbalances present in both urban and sub-urban environments. For example, the total points captured from a typical street scene using a Terrestrial Laser Scanner (TLS) or Mobile Laser Scanner (MLS) can consist of $>$90\% road and fa{\c c}ade points. Similarly, features such as pole-like objects, pedestrians and street furniture contain few points due to their comparatively small size and natural scarcity of occurrence. This issue has been well acknowledged within machine learning for point cloud classification \cite{WeinmannEtAlM2015}, however, typically, these have been for classical machine learning approaches such as support vector machines and random forests. These methods learn on individual points and therefore have abundant training data. The solution to balance classes is therefore to reduce the number of examples for strongly represented classes to the quantity of least represented classes, that meet a certain minimum threshold. Such a method is not sufficient for training Deep Neural Networks (DNNs) as modern DNNs operate on batches of points and thus need much larger data sets to achieve high classification accuracy. It is therefore unfavourable to reduce the point cloud as in some cases this would result in rejecting $> 90$\% of points in the training dataset.

The current best-practise to account for class-imbalance with training DNNs is to scale the networks loss based on a per-class weight coefficient. This helps to prevent under-represented classes being over-shadowed by abundant classes. The weight coefficients can be determined as a function relating to the probability of the point occurring in a scene. In this paper, we propose an additional pre-processing stage to help further address this issue, by physically reducing the class im-balance through selective augmentations. Our approach quantifies the representation of a given scene by analysing it's class occurrences. Scenes containing many points which are under-represented score higher. The number of augmentations is then determined as a non-linear function of the derived score. By weighting the augmentations in this way the class balance is subsequently reduced. We test our hypothesis using the popular PointNet++ network architecture on both indoor (ScanNet) and outdoor (Semantic3D) datasets.

\section{Related work}

Within the field of deep learning there has been an active effort to address the issue of class-imbalance. The most common approach is to weight the loss function with the inverse frequency of the labels occurrence. This method was proposed by \cite{lin2017focal} to address the class-imbalance in object detection CNN's where in many cases the majority of classifications are easy to detect background. This is achieved by re-shaping the cross-entropy by adding a modulating factor, ensuring negative/frequent classes do not overwhelm the loss function. This was shown to improve the performance for single-class object detectors \cite{GriffithsBoehmD2018}, where class-imbalance is likely to be high. The ability to pass weights into cross-entropy loss function is now a standard feature for many leading deep learning software libraries. \cite{YueS2017} proposed a method by which the softmax loss function is scaled by a scaling parameter determined as a function of the labels frequency. In essence, this is a reactive approach for dealing with scenarios where class im-balance is assumed. \cite{FidonEtAlL2018} propose using a generalised \textit{Wasserstein Dice Score} to take advantage of inter-class relationships and multi-scale information. The improved loss function favours semantically meaningful predictions, which can help balance mis-classification due to class im-balance.

Alternative approaches include undersampling and oversampling data. Undersampling is the process of randomly removing data from classes which are highly represented such that their ratio approaches that of the under-represented classes e.g. \cite{WeinmannEtAlM2015}. In contrast, oversampling is the procedure of replicating under-represented classes such that they approach the count of highly-represented classes. Whilst undersampling can result in a lot of useful information invariably being lost, oversampling results in replication of identical data which can also lead to over-fitting to small data samples. A more sophisticated approach is proposed by \cite{ChawlaEtAlN2002} in their technique called Synthetic Minority Over-sampling (SMOTE). SMOTE suggests a combination of undersampling and oversampling is the most effective approach. Oversampling is further achieved by generating synthetic data that is similar to the original under-represented class data when assessed using a nearest-neighbour classification. Although this method demonstrated promising results, it has not been tested in the context of deep learning. The authors also tested the possibility of bagging and boosting methods, however, these methods were not shown to improve the performance of the classification algorithm. Nevertheless, none of these approaches address the issue when there is an extremely under-represented class. As such, there is still no common heuristic for dealing with class-imbalance with under/oversampling. The work presented here follows a similar approach of synthetic oversampling, however, we use augmentation to generate the synthetic data.

More recently, \cite{ZhuEtAlX2017} have proposed the use of Generative Adverserial Networks (GAN) to generate data from the \textit{true} distribution of the data. By supplementing the data manifold with an approximation from the \textit{true} distribution, classification rates were shown to improve by 5\%-10\% in emotion classification for 2D images. Whilst this method shows promise, the use of GANs for 3D point cloud generation is still in its infancy \cite{AchlioptasEtAlP2017}, and not currently capable of generating convincing data matching the complexity of ScanNet/Semantic3D.

\section{Methodology}

To evaluate our proposed weighted augmentation approach we train and validate our results on two common datasets; ScanNet and Semantic3D. Each dataset consists of numerous point cloud scenes where a point cloud is defined as a set of 3D points ${P_{i}|i=1,...,n}$ where $P \in \mathbf{R}^{3}$ such that $P_{i}$ is a vector ($x,y,z$) denoting its location in a euclidean coordinate system. We select ScanNet and Semantic3D for two main reasons. Firstly, both datasets contain a large number of points (750m and 4bn respectively) necessary for training deep neural networks, and have demonstrated themselves as standard benchmark datasets for deep learning with point clouds. Secondly, each dataset contains a varied class im-balance, with the indoor ScanNet having a more even class distribution to the outdoor Semantic3D. This allowed us to evaluate the effect of weighted augmentations over varying class-imbalances. Each dataset was pre-processed using the same pipeline which we describe below.

\subsection{Preprocessing}

The initial stage to pre-processing was to determine the label weights for each dataset. A general heuristic for calculating class weights for point clouds is defined by \cite{DaiEtAlA2017} as $1/log(1.2+\mbox{probability of occurence})$. After internal experiments we opted for normalised weights between $0-1$ which a capped lower threshold ($t_{min}$). We compute our weights ($w$) as:

\begin{equation}
w = (t_{max}-t_{min})(\frac{-\sum_{i=1}^{n}[P_{i}=x]-\min{P}}{\max{P}-\min{P}})+t_{min}
\end{equation}

\noindent where $t_{max}$ and $t_{min}$ are the maximum and minimum weight thresholds respectively, $p$ is the entire point set and $x$ is the class for which the weight is to assigned.

By scaling the weights between a minimum and maximum threshold we increase the variance of weights at the upper end of the distribution, and retain the ability to cap a minimum threshold. In our experiments we found $t_{min}=0.25$ yielded the best results for both datasets. 

To feed the data into a PointNet++ network the entire dataset needs to be split into chunks of point clouds of uniform size $n$, where $n$ is equal to the number of input nodes for the network (8192 in our case). To achieve this we first split the dataset into a planar grid using a 10x10m grid size. As the point clouds have varying point density, each chunk ($C$) contains an undefined number of points ($C_{n}$). The simplest approach to deal with this is to discard any chunks where $C_{n}<n$, and randomly sub-sample $C$ where $C_{n}>n$. To improve on scenarios where $C_{n}>n$, we incorporate an adaptive voxel down-sampling approach. The initial voxel size $v_{b}$ where b is a cube, is set to $0.01^{3}$m for ScanNet and $0.05^{3}$m for Semantic3D. All points that fall within the voxel are represented by a new point $p_{i}$ with the coordinates of the voxel centroid, and the label is determined via a maximum vote scheme. While $C_{n}>n$ we incrementally increase the value $b$ on the original chunk point cloud until $C_{n}<n$, we then take the value of $b$ where $C_{n}$ is as close as possible to $n$, but larger \ref{variable_voxel}. This ensures the points are primarily reduced in a geometrically and spatially coherent manner, before employing probabilistic sub-sampling to achieve the exact target number. To initially help reduce class im-balance points are sub-sampled from a non-uniform distribution where the probability $Pr$ of a point being sub-sampled is the inverse of the corresponding class weight $w$ such that $Pr=f(-w)$. Finally, to avoid discarding too many valid points where $C_{n}<n$, if $C_{n} >= 0.5n$ we randomly duplicate points until $C_{n}=n$.

\begin{algorithm}
	
	v$=$0.01\\
	large chunk = reduced chunk = original chunk\\
	\While{size(reduced chunk) $>$ n}{
	    large chunk = reduced chunk\\
		reduced chunk = voxel-downsample(original chunk, v) \\
		v += increment
	}
	final chunk = large chunk
	\caption{Adaptive voxel downsampling}
	\label{variable_voxel}
	
\end{algorithm}

We split the dataset into training (60\%), test (20\%) and validation (20\%) batches. The test sub-set is used for in-training performance evaluation. The model state which achieves the highest performance on the test data is subsequently exported and then used for inference on the validation sub-set which gives the final performance of the model (\textit{Section \ref{results_section}}).

\subsection{Augmentation}

We define an augmentation as a random rotation in the $x$ and $y$ axis about the $z$ axis, followed by a small rotation in the $x, y, z$ axis'. We refer to this as a single permutation. To apply the permutation we randomly compute a value $r$ and multiply by rotation matrix $R$ such that:

\begin{gather}
\mathit{p=}
\begin{bmatrix*}[c]
\cos{2r\pi} & -\sin{2r\pi} & 0 \\
\sin{2r\pi} & \cos{2r\pi}  & 0 \\
0 & 0 & 1
\end{bmatrix*}
\begin{bmatrix*}[c]
r_{11} & r_{12} & r_{13} \\
r_{21} & r_{22} & r_{23} \\
r_{31} & r_{32} & r_{33}
\end{bmatrix*}
\end{gather}
\noindent where $r$ is a randomly generated number such that $0<r<1$.

To determine the number of augmentations $a_{n}$ we quantify the chunk based to the number of under-represented classes present within the chunk where $a_{n}=f(\sum_{i=1}^{n}[P_{i}=x], w)$ for $x \cap c$. We call this value the chunk uniqueness ($u$). We calculate $u$ by first normalising the counts of each class present in the chunk between by $c_{n}$. Then we take the sum of all of the associated normalised counts for each label in the chunk and multiply by $w$. This returns a quantified value such that $u=[0, 1]$. Formally we define this as:

\begin{equation}
u = \frac{\sum_{i=1}^{k}\sum_{j=1}^{n}[P_{j}=x_{i}]}{n} * w_{k}
\end{equation}
\noindent where $k$ is the number of classes in the chunk.

Finally to determine the $a_{n}$ we calculate:

\begin{equation}
a_{n} = \frac{10\tan{u}^{2}}{2}
\end{equation}

This yields the following values for $a_{n}$:

\small
\begin{tabular}{lrrrrrrr}
	\hline
	$u$ & 0.25 & 0.375 & 0.5 & 0.625 & 0.75 & 0.875 & 1 \\
	$a_{n}$ & 1 & 1 & 2 & 3 & 5 & 8 & 13 \\
	\hline
\end{tabular}
\normalfont

\begin{figure}[H]
	\centering
	
	\subfloat[ScanNet]{
		\label{subfig:scannet_class_dist}
		\includegraphics[width=0.45\textwidth]{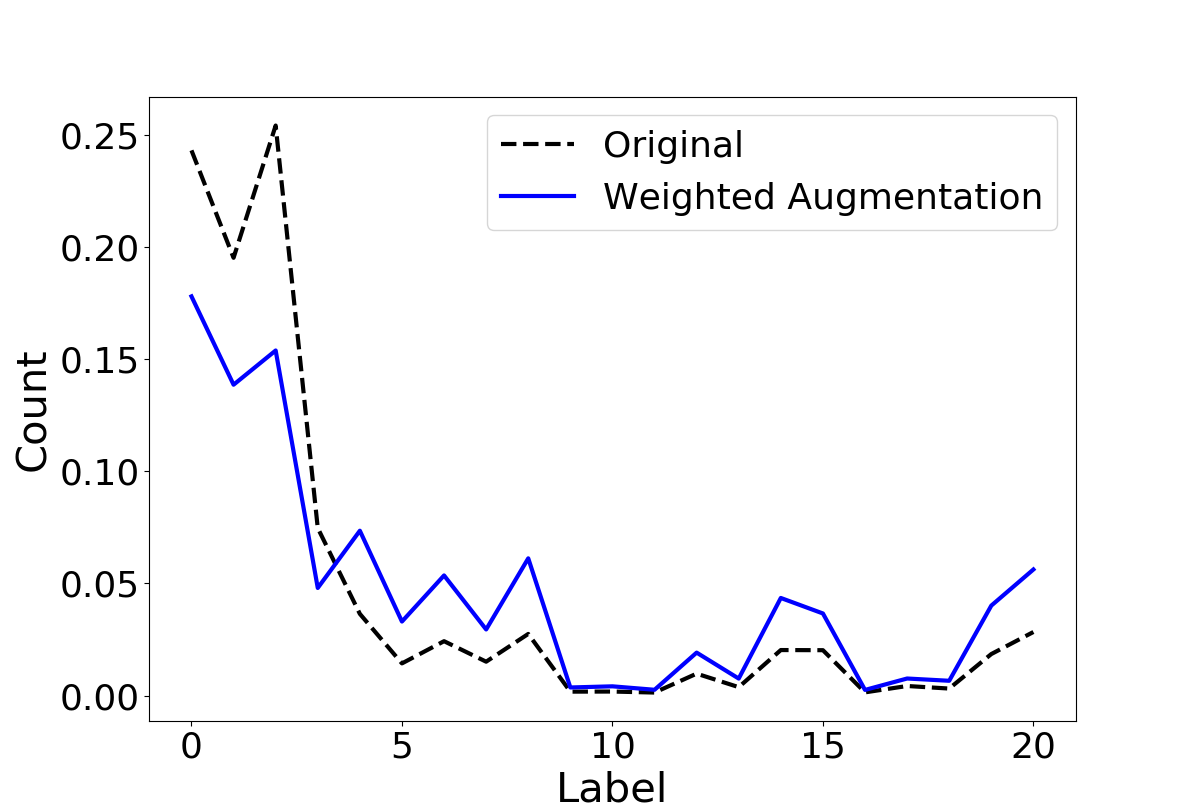}}
	
	\subfloat[Semantic3D]{
		\label{subfig:semantic3d_class_dist}
		\includegraphics[width=0.45\textwidth]{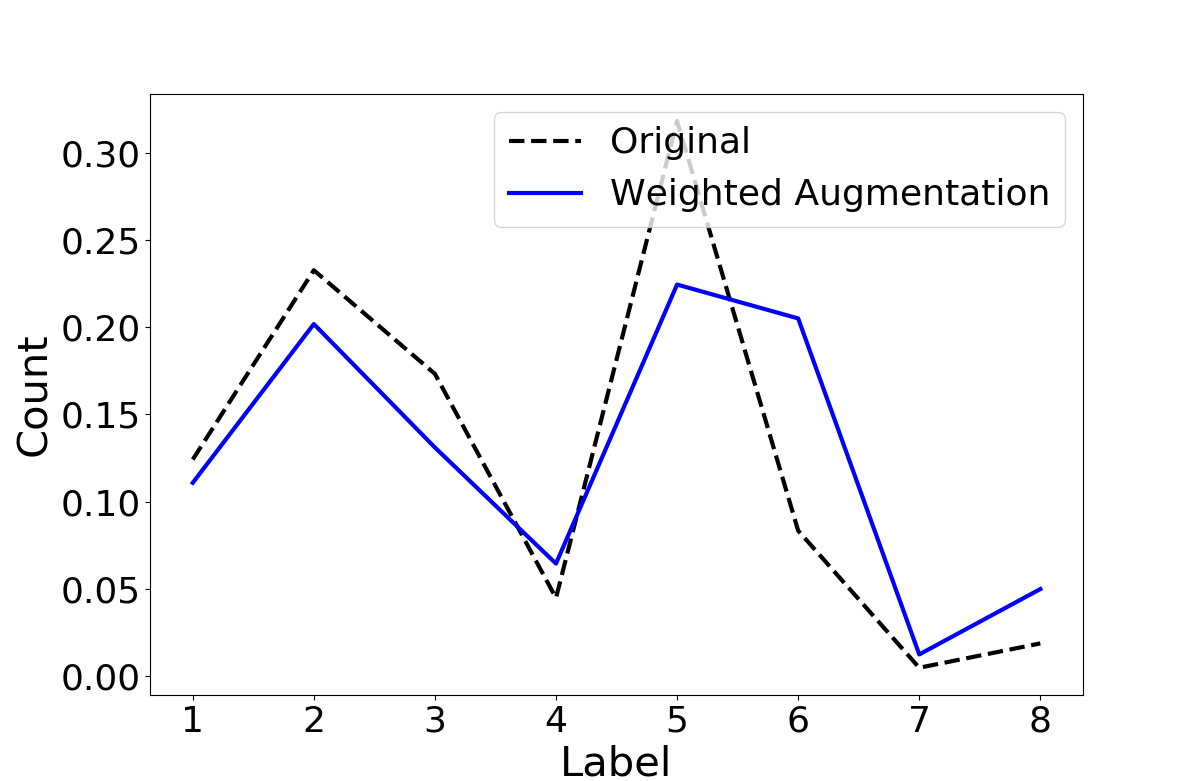}}
	
	\caption{Normalised distribution of classes of datasets before and after weighted augmentation and non-uniform sampling for a) ScanNet and b) Semantic3D. A more horizontal line indicates  }
	\label{fig:class_distributions}
\end{figure}

By scaling $a_{n}$ by $tan^{2}$ this ensures that highly under-represented chunks are augmented substantially more than higher-represented scenes. The affect this procedure has on the class distribution of the datasets can be seen visually in Figure \ref{fig:class_distributions}.

\subsection{Network architecture}

We employ PointNet++ for end-to-end model training. PointNet++ is an extension of the seminal deep learning point cloud architecture PointNet \cite{CharlesEtAlR2017}. Unlike previous deep learning approaches for end-to-end 3D point cloud processing, PointNet does not extract features with a 3D convolution operator (i.e. \cite{Maturana2015}), but instead consists only of fully-connected layers. Features are generated using Multi Layer Perceptrons (MLPs) and aggregated using a single \textit{symmetric function}, max-pooling. In essence, the network learns a set of functions that select interesting and informative key-points from a sub-set of points, encoding this information in each layers feature vector. Semantic segmentation is achieved by concatenating the aggregated global features into two MLPs to generate per-point features and subsequently class probabilities for each point. Per-point features are obtained by concatenating global feature vectors with each of the point features.

PointNet++ is a hierarchical network extension of PointNet which has the ability to capture local structures induced by the metric space points live in. Point sets are partitioned into overlapping local regions by a distance metric. Features are then extracted from progressively increasing neighbourhood sizes. Whereas small neighbourhoods capture fine-grain local features (i.e. surface texture), large neighbourhoods capture global shape geometry features. Overlapping partitions are generated with a neighbourhood ball with centre $p_{x,y,z}$ and radius $r$. Where $p$ is each point the set.

Model hyper-parameters are initially derived from the original values outlined by \cite{QiEtAlC2017a}, with a few minor changes. To reduce result ambiguity and retain focus on the affects of weighted augmentation, these values are not further revised for each training scenario. The final model hyper-parameters were; batch size = 16, learning rate = 0.001, momentum = 0.9, weight decay rate = 0.7, number of input points (single batch) = 8192. We did not make any changes to the network architecture, and therefore the reader is referred to the original paper for further details. Each network was trained for 25 epochs on a single Nvidia GTX 1080 Ti graphics processing unit which took between ~15-30 hours and ~20-40 hours for Semantic3D and ScanNet datasets respectively.

\subsection{Performance evaluation}

We evaluate performance with 5 metrics for each processing scenario. These are; precision, recall, F1, accuracy and mean intersection over union (IoU). We define these as; precision$=\frac{tp}{tp+fp}$, recall$=\frac{tp}{tp+fn}$ and F1$=2\frac{recall*precision}{recall+precision}$ where $t, f, p, n$ are true, false, positive and negative respectively. Mean IoU is calculated from the confusion matrix where intersection $i$ is the intersection of correct predictions (diagonal top left to bottom right). Union $u$ is the sum of both the predicted and true label columns for each class respectively. Mean IoU is then calculated as $\frac{\sum_{j=1}^{i=k}u_{j}-i_{j}}{k}$, where $k$ is the number of classes. 

\begin{table*}[t]
	\centering
	\caption{Semantic3D and ScanNet validation dataset performance results. In all scenarios augmentation performed better than when no augmentation was used. Weighted augmentation further increases performance.} \label{tab:results}
	\begin{tabular}{lrrrrr}
		
		\textbf{Training scenario} & \textbf{Precision} & \textbf{Recall}& \textbf{F1} & \textbf{mean IoU} & \textbf{Accuracy (\%)} \\
		\hline
		
		Semantic3D vanilla & 0.466 & 0.440 & 0.443 & 0.938 & 87.2 \\
		Semantic3D augmentation & 0.497 & 0.475 & 0.478 & 0.950 & 89.7 \\
		Semantic3D weighted augmentation & \textbf{0.564} & \textbf{0.552} & \textbf{0.554} & \textbf{0.956} & \textbf{98.1} \\
		
		\hline
		
		ScanNet vanilla & 0.716 & 0.705 & 0.696 & 0.780 & 89.4 \\
		ScanNet augmentation & 0.779 & 0.779 & 0.765 & 0.791 & 90.4 \\
		ScanNet weighted augmentation & \textbf{0.841} & \textbf{0.848} & \textbf{0.835} & \textbf{0.842} & \textbf{93.2} \\
		
		\hline
		
	\end{tabular}
\end{table*}

\section{Results}\label{results_section}

\begin{table}[h]
    \small
    \centering
	\caption{ScanNet normalised confusion matrix intersection (\textbf{I}) values for no augmentations (a) and weighted augmentations (b) processing scenarios.} \label{tab:scannet_conf}
	\begin{tabular}{lccccccc}
        \hline
		\textbf{Class} & \textbf{0} & \textbf{1} & \textbf{2} & \textbf{3} & \textbf{4} & \textbf{5} & \textbf{6} \\
		\textbf{I} & 0.70 & 0.88 & 0.97 & 0.97 & 0.95 & 0.95 & 0.99 \\
		\hline
		 & \textbf{7} & \textbf{8} & \textbf{9} & \textbf{10} & \textbf{11} & \textbf{12} & \textbf{13} \\
		 & 0.96 & 0.98 & 0.87 & 0.97 & 0.98 & 0.97 & 0.90 \\
		\hline
	     & \textbf{14} & \textbf{15} & \textbf{16} & \textbf{17} & \textbf{18} & \textbf{19} & \textbf{20} \\
	     & 0.91 & 0.94 & 0.96 & 0.94 & 0.85 & 0.97 & 0.93 \\
		\hline
		\\
		\\
        \hline
		\textbf{Class} & \textbf{0} & \textbf{1} & \textbf{2} & \textbf{3} & \textbf{4} & \textbf{5} & \textbf{6} \\
		\textbf{I} & 0.81 & 0.95 & 0.97 & 0.98 & 0.98 & 0.98 & 0.99 \\
		\hline
		 & \textbf{7} & \textbf{8} & \textbf{9} & \textbf{10} & \textbf{11} & \textbf{12} & \textbf{13} \\
		 & 0.98 & 0.99 & 0.95 & 0.98 & 0.99 & 0.99 & 0.93 \\
		\hline
	     & \textbf{14} & \textbf{15} & \textbf{16} & \textbf{17} & \textbf{18} & \textbf{19} & \textbf{20} \\
	     & 0.98 & 0.98 & 0.99 & 0.99 & 0.97 & 0.98 & 0.98 \\
		\hline
	\end{tabular}
\end{table}
\normalfont

\begin{figure}[t]

	\subfloat[]{
		\label{subfig:1}
		\includegraphics[width=0.46\textwidth]{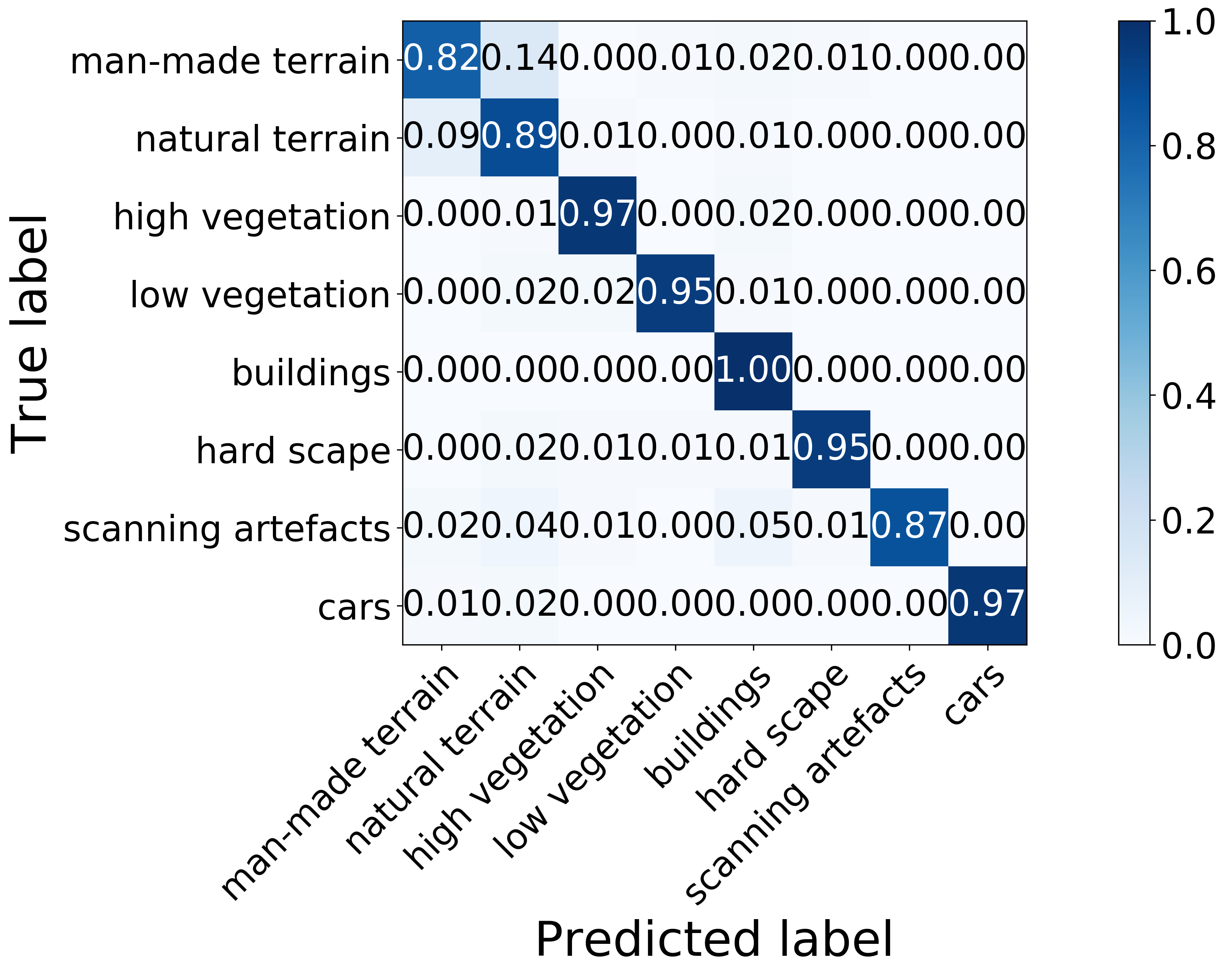}}
	
	\subfloat[]{
		\label{subfig:2}
		\includegraphics[width=0.46\textwidth]{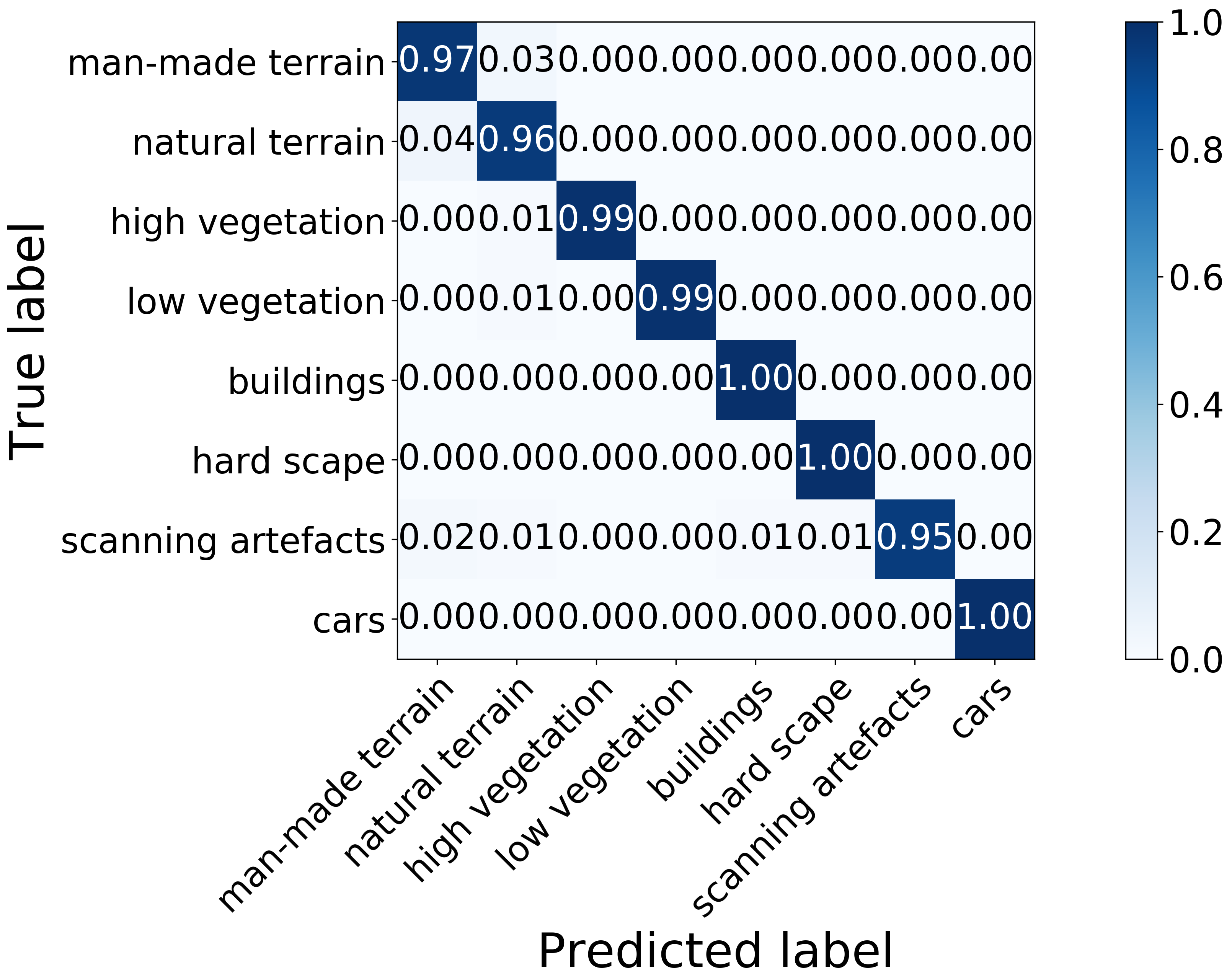}}

	\caption{Confusion matrices for a) Semantic3D with no augmentation and b) Semantic3D with weighted augmentation.}
	\label{fig:confusion_mat}
\end{figure}

The results for each experiment are presented in Table \ref{tab:results}. In both Semantic3D and ScanNet datasets, models trained with some form of augmentation resulted in higher validation scores. This was more prominent with respect to overall accuracy on the outdoor Semantic3D dataset where a higher class-imbalance existed both before and after pre-processing. Caution should be sought when measuring success with overall accuracy when any form of class im-balance is present as this can indicate over-fitting on the dominant class, resulting in a model that scores highly but generalises across classes poorly. Despite this, gains were also made with respect to precision and recall values for both datasets, suggesting multi-class improvements. This is further justified by the confusion matrices seen in Figure \ref{fig:confusion_mat} and Table \ref{tab:scannet_conf}. In each scenario the incorporation of class im-balance reduction led to not only an increase in correct classifications of poorly represented classes, but also for highly represented classes. This suggests the model is generalising better than when class-imbalance is lower, which suggests that the network is benefiting from a more diverse dataset. The very high increase in overall accuracy experienced on the Semantic3D dataset, is likely due to reduction of incorrect classifications of man-made terrain and natural terrain as they are both dominant classes and the accuracy is not a weighted value.

\begin{figure}[H]

	\subfloat[]{
		\label{subfig:semantic3d_classification}
		\includegraphics[width=0.47\textwidth]{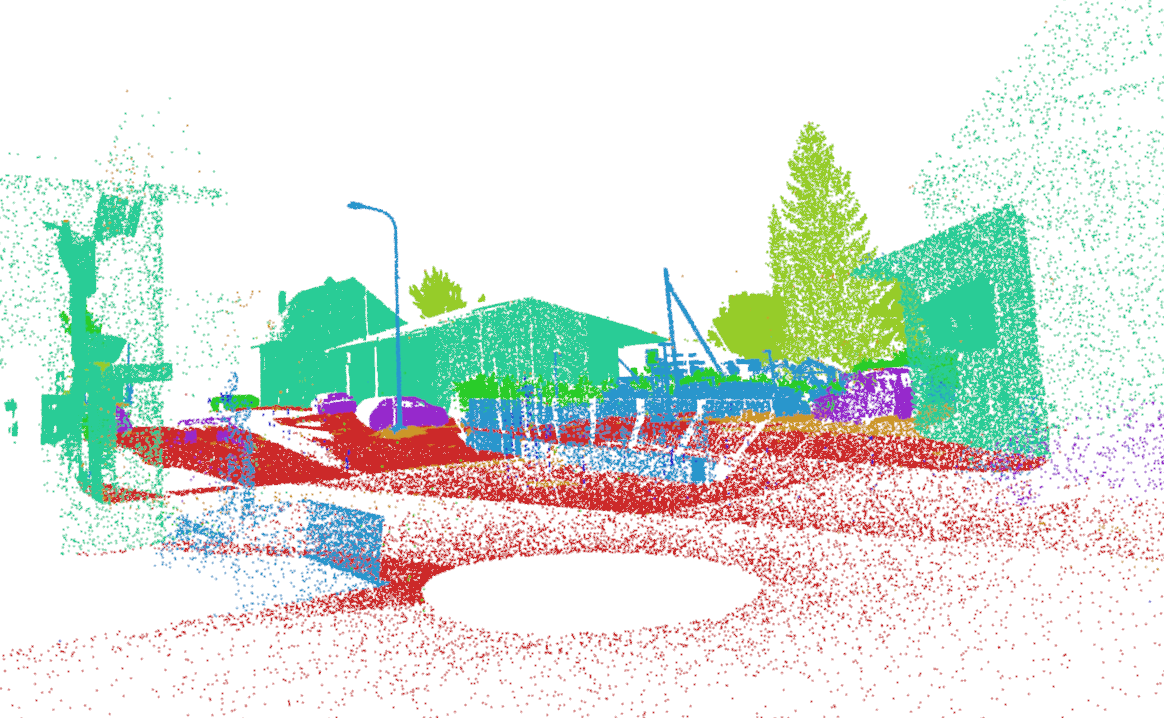}}

	\subfloat[]{
		\label{subfig:scannet_classification}
		\includegraphics[width=0.47\textwidth]{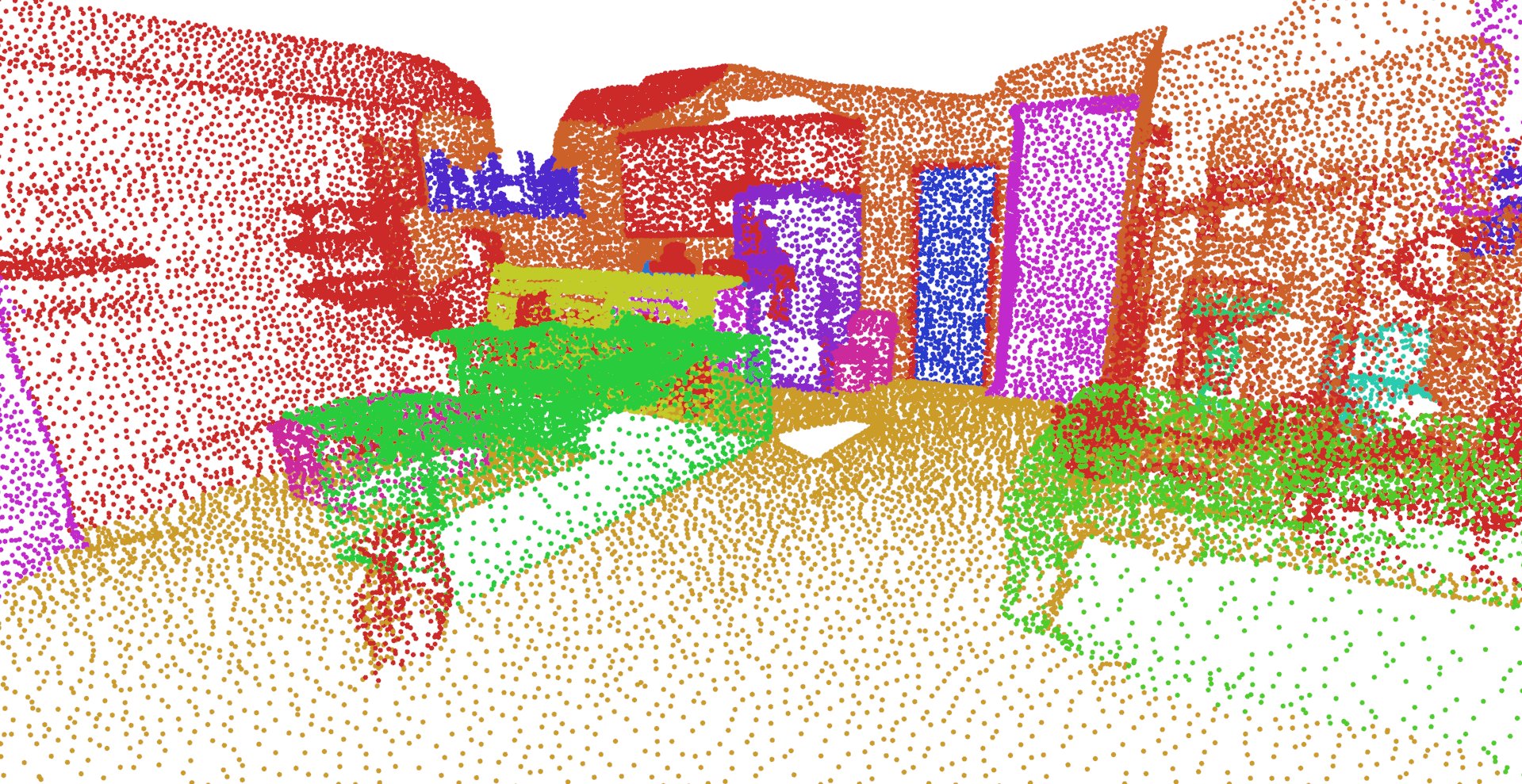}}

	\caption{Point cloud classification results of a) Semantic3D and b) ScanNet datasets. Images are derived from inference of entire scene, which contains training, test and validation examples.}
	\label{fig:sematic3d_pointcloud}
\end{figure}

Interestingly, there appears to be no correlation with respect to the confusion matrix intersection score improvement and the number of samples per-class. Again, this also suggests that the model is generally performing better over all classes. A concern would be if poorly represented class classification accuracy was improved at the cost of highly represented classes, however in our experiments this does not appear to be the case.

In each scenario, the gain from vanilla to augmentation was less than the gain from augmentation to weighted augmentation, demonstrating the advantages of such a strategy. For example, Semantic3D F1 scores increased 7.9\% from vanilla to augmentation and 15.9\% from augmentation to weighted augmentation. Similarly, with respect to overall accuracy 2.5\% was gained from augmentation, but from augmentation to weighted augmentation 8.4\% was achieved. ScanNet also had similar conclusions with F1 increases of 8.8\% and 9.6\% respectively, and for overall accuracy 1.4\% and 4\% respectively.

\section{Discussion}

The results discussed in Section \ref{results_section} suggest a overall improvement was witnessed by the incorporation of class-imbalance reducing procedures, namely, weighted augmentation along with non-random sub-sampling. Whilst these results are promising, we would argue that their conclusions should still be taken cautiously. For example, the validation results of Semantic3D out perform the current benchmark leaders, however, this was not evaluated on the full test set. Validation chunks were taken from within the same scenes that both training and test data came from. Furthermore, these values are taken from a sparse classification of the point cloud as apposed to a full point classification. To fully classify a point cloud further steps must be taken such as a K-nearest-neighbour and interpolation to achieve a dense classification. The purpose of this experiment was solely to determine if confusion matrix ambiguity could be minimised by reducing the class imbalance. The results never-the-less demonstrated potential for overall improvements and further work should look into validation on new datasets.

Analysis of the confusion matrix and results combined strongly indicates the importance of class-balance for training robust DNNs on geometric data. This was most prominent by the improvement in performace on dominant classes after the increase in number of points for less dominant classes. This suggests that models that achieve high performance scores where high class im-balance is occurring are subject to some forms of class-specific over-fitting, in which the results presented in this paper suggest is worse performing than a model with more balanced classes, especially where the classes are geometrically variant. It is still unlikely that this has been completely mitigated from either dataset, in particular Semantic3D where an initially higher class im-balance was present. Evidence of this can be observed in the discrepancy between overall accuracy and F1 scores for each dataset. Whereas Semantic3D has a higher overall accuracy the F1 score is substantially lower when compared to ScanNet. Mitigation of class im-balance should hopefully address this issue by minimising this discrepancy, ideally by raising the precision and recall values. It would therefore seem reasonable to assume in datasets where overall accuracy is significantly higher than F1 score, class im-balance could be an influencing factor. Furthermore, this suggests that an ideal training dataset for point cloud classification with DNNs is both geometrically balanced, as well as class balanced with respect to total counts of points.

Although we limit the learning to features derived purely from each point's $x,y,z$ components, due to the connectionist nature of neural networks, it remains difficult to conclude without the need for proxy indicators, what features the network has learned. Voxel down-sampling was performed to ideally remove the potential for the network to learn point density, however, it is still not possible to conclude that the networks features are purely derived from geometry. So although the reduction of class im-balance led to an overall improvement across all classes, it is not possible to accredit this to more geometrically meaningful features.

\section{Conclusion}

In this paper we present weighted augmentations as a pre-processing technique for training DNNs where large class im-balances occur within the training data. A normalised weighting function was described to derive individual class weights dependant on the probability of occurrence within the training data. Individual geo-spatially chunked point cloud sets are then assigned a quantifiable metric to determine the uniqueness of the chunk, based on the presence of points with highly weighted classes. From this scene uniqueness metric we compute a value $a_{n}$ from a non-linear function, where $a_{n}$ is the number of augmentations applied to the chunk. By strongly augmenting scenes with many under-represented classes we reduce the total class im-balance present in the training data. We further address the class im-balance by using the class weights to derive the probability of selection in a non-uniform sub-sample when the chunk contains more points than input nodes of the DNN. Experiments undertaken with the ScanNet and Semantic3D datasets using the PointNet++ architecture suggest that reduction of the class-imbalance has a positive influence on the performance of the network. An increase in F1 score of 19\% and 25\% and overall accuracy value of 3.8\% 10.9\% for ScanNet and Semantic3D respectively was observed when weighted augmentations were used to reduce the class im-balance. These results suggest that the reduction of class im-balance can have a significant affect on model training, especially when the im-balance is very strong, for example in outdoor environments.

\bibliography{references}

\vspace{1cm}
\textit{Revised April 2019}

\end{document}